\pdfoutput=1

\documentclass{article}
\usepackage[accepted,nohyperref]{icml2016}
\usepackage{times}
\usepackage{graphicx} 
\usepackage{subfig} 

\usepackage{amssymb}
\usepackage{setspace}

\icmltitlerunning{Maximum Entropy Deep Inverse Reinforcement Learning}

\newcommand{\fis}{1.45 in}
\newcommand{\fisb}{0.8 in}

\definecolor{myblue}{RGB}{0,51,102}

\begin{document} 

\twocolumn[
\icmltitle{Maximum Entropy Deep Inverse Reinforcement Learning}

\icmlauthor{Markus Wulfmeier}{markus@robots.ox.ac.uk}
\icmlauthor{Peter Ondr\'{u}\v{s}ka}{ondruska@robots.ox.ac.uk}
\icmlauthor{Ingmar Posner}{ingmar@robots.ox.ac.uk}
\icmladdress{Mobile Robotics Group, 
	Department of Engineering Science, 
	University of Oxford}

\icmlkeywords{Inverse Reinforcement Learning, Neural Networks, Deep Learning, Maximum Entropy, Feature Learning}

\vskip 0.3in
]

\begin{abstract} 
This paper presents a general framework for exploiting the representational capacity 
of neural networks to approximate complex, nonlinear reward functions in the context of solving the inverse reinforcement learning (IRL) problem.
We show in this context that the Maximum Entropy paradigm for IRL lends itself naturally to the efficient training of deep architectures.
At test time, the approach leads to a computational complexity independent of the number of demonstrations, which makes it especially well-suited for applications in life-long learning scenarios.
Our approach achieves performance commensurate to the state-of-the-art on existing benchmarks while exceeding on an alternative benchmark based on highly varying reward structures.
Finally, we extend the basic architecture - which is equivalent to a simplified subclass of Fully Convolutional Neural Networks (FCNNs) with width one - to include larger convolutions in order to eliminate dependency on precomputed spatial features and work on raw input representations.
\end{abstract}

\section{Introduction}
\label{introduction}
Recent successes in machine learning, vision and robotics have lead to widespread expectations that machines will increasingly succeed in applications of real value to the public domain. A central tenet of any vision delivering on this promise revolves around learning from user interactions. 
Inverse reinforcement learning (IRL) is playing a pivotal role in these developments and commonly finds applications in robotics \cite{argall2009survey} where it allows robot to learn complex behaviour from human demonstrations and also in fields of cognition \cite{baker2009action} and preference learning \cite{ziebart2008maximum} where it serves as a tool to better understand human decisions or medicine \cite{asoh2013application} to predict patient response to treatment.
The objective of inverse reinforcement learning (IRL) is to infer the underlying reward structure guiding an agent's behaviour based on observations as well as a model of the environment. This may be done either to learn the reward structure for modelling purposes or to provide a method to allow the agent to imitate a demonstrator's specific behaviour ~\cite{ramachandran2007bayesian}. 
While for small problems the complete set of rewards can be learned explicitly, many problems of realistic size require the application of generalisable function approximations.

\begin{figure}
	\centering
	\includegraphics[height=2.4in]{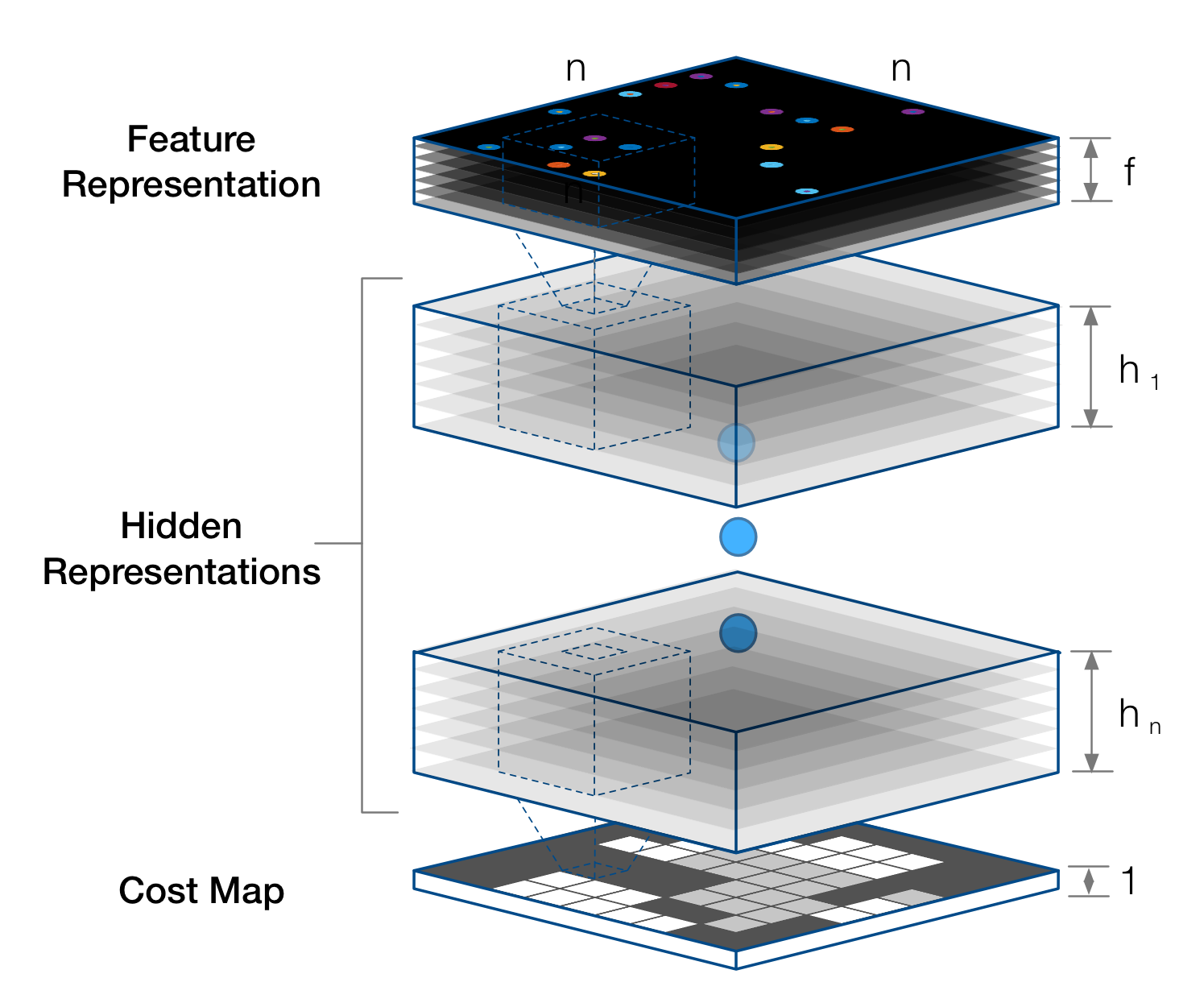}
	\caption{Fully Convolutional Neural Network for reward approximation in the IRL setting. The network serves to model the relationship between input features and final reward map. }
	\label{fig:fcnn}
\end{figure}

Much of the prior art in this domain relies on parametrisation of the reward function based on pre-determined features. In addition to better generalisation performance than direct state-to-reward mapping, this approach enables the transfer of learned reward functions between different scenarios with the same feature representation. A number of early works from \cite{ziebart2008maximum}, \cite{abbeel2004apprenticeship}, \cite{lopes2009active} and \cite{ratliff2006maximum}, express the reward function as a weighted linear combination of hand selected features. To overcome the inherent limitations of linear models, \cite{choi2013bayesian} and \cite{levine2010feature} extend this approach to a limited set of nonlinear rewards by learning a set of composites of logical conjunctions of atomic features. Non-parametric methods such as Gaussian Processes (GPs) have also been employed to cater for potentially complex, nonlinear reward functions \cite{levine2011nonlinear}. While in principle this extends the IRL paradigm to the flexibility of nonlinear reward approximation, the use of a kernel machine makes this approach scale badly with higher numbers of training data and prone to requiring a large number of reward samples in order to approximate highly varying reward functions \cite{bengio2007scaling}. 
Even sparse GP approximations as used in \cite{levine2011nonlinear} lead to a query complexity time in dependency of the size of the active set or the number of experienced state-reward pairs.
Situations with increasingly complex reward function leading to higher requirements regarding the number of inducing points can quickly render this nonparametric approach computationally impracticable.
Furthermore, in comparison to \cite{babes2011apprenticeship}, we focus on a singular expert in what finally leads to an an end-to-end learning scenario in section \ref{sec:cnn} from raw input to reward without compression or preprocessing on the input representation.
To our knowledge the only other work considering the use of deep networks is given by \cite{levine2015rss}, who focus on directly approximating policies with neural networks but shortly refer to the possibility of extension for cost function learning with neural networks.

In contrast to prior art, we explore the use of neural networks to approximate the reward function. Neural Networks already achieve state-of-the-art performance across a variety of domains such as computer vision, natural language processing, speech recognition \cite{bengio2012review} and reinforcement learning \cite{mnih2013playing}. 
Their application in IRL suggests itself due to their compact representation of highly nonlinear functions through the composition and reuse of the results of many nonlinearities in the layered structure \cite{bengio2007scaling}. 
In addition, NNs provide favourable computational complexity ($\mathcal{O}(1)$) at query time with respect to observed demonstrations, which provides for scaling to problems with large state spaces and complex reward structures -- circumstances which might render the application of existing prior methods intractable or ineffective.
With the approach represented in Figure \ref{fig:fcnn}, a state's reward can be determined either solely based on its own feature representation or -- in using wider convolutional layers -- analysed in combination with its spatial context. The applied architectures are Fully Convolutional Neural Networks, which -- by skipping the fully connected final layers common in classification tasks -- preserve spatial information and can create an output of the same spatial dimension and size as the input. Recent examples for the application of FCNNs focus on dense prediction: including pixel-wise semantic segmentation by \cite{LongSD14}, sliding window detection and prediction of object boundaries \cite{SermanetEZMFL13}, depth estimation with single monocular images \cite{LiuSLR15} and human pose estimation in monocular images \cite{TompsonJLB14}.

Our principal contribution is a framework for \emph{ Maximum Entropy Deep Inverse Reinforcement Learning} (DeepIRL) based on the Maximum Entropy paradigm for IRL \cite{ziebart2008maximum}, which lends itself naturally for training deep architectures by leading to an objective that is - without approximations - fully differentiable with respect to the network weights. Furthermore, we demonstrate performance commensurate to state-of-the-art methods on a publicly available benchmark, while outperforming the state-of-the-art on a new benchmark where the true underlying reward has complex interacting structure over the feature representation. 
In addition, we emphasise the flexibility of the approach and eliminate the requirement of preprocessing and precomputed features by applying wider convolutional layers to learn spatial features of relevance to the IRL task. This enables the application without manually crafted feature design as long as the state space is constrained to a regularly gridded representation allowing for convolutions.

We argue that these properties are important for practical large-scale applications of IRL as can be seen in life-long learning approaches with often complex reward functions and increasing scale of demonstrations requiring high capacity models and fast computational speeds.

\section{Inverse Reinforcement Learning}
This section presents a brief overview of IRL. 
Let a Markov Decision Process (MDP) be defined as $\mathcal{M} = \{ \mathcal{S}, \mathcal{A}, \mathcal{T}, r \}$, where $\mathcal{S}$ denotes the state space, $\mathcal{A}$ denotes the set of possible actions, $\mathcal{T}$ denotes the transition model and $r$ denotes the reward structure. Given an MDP, an optimal policy $\pi^*$ is one which, when adhered to, maximizes the expected cumulative reward. Furthermore, an additional factor $\gamma \in [0,1]$ may be considered in order to discount future rewards.

IRL considers the case where a MDP specification is available but the reward structure is unknown. Instead, a set of expert demonstrations $\mathcal{D} = \{ \varsigma_1, \varsigma_2, ..., \varsigma_N \}$ is provided which are sampled from a user policy $\pi$, i.e. provided by a demonstrator. Each demonstration consists of a set of state-action pairs such that $\varsigma_i = \{(s_0,a_0),(s_1,a_1),...,(s_K,a_K)\}$. The goal of IRL is to uncover the hidden reward $r$ from the demonstrations. 

A number of approaches have been proposed to tackle the IRL problem (see, for example, \cite{abbeel2004apprenticeship}, \cite{neu2012apprenticeship}, 
\cite{ratliff2006maximum}, \cite{syed2007game}). An increasingly popular formulation is Maximum Entropy IRL \cite{ziebart2008maximum}, which was used to effectively model large-scale user driving behaviour. In this formulation the probability of user preference for any given trajectory between specified start and goal states is proportional to the exponential of the reward along the path
\begin{equation}
P(\varsigma|r) \propto \exp \{ \sum_{{s,a} \in \varsigma} r_{s,a} \}.
\label{eq:trajectoryLikelihood}
\end{equation}

As shown in Ziebart's work, principal benefits of the Maximum Entropy paradigm include the ability to handle expert suboptimality as well as stochasticity by operating on the distribution over possible trajectories.
Moreover, the Maximum Entropy based objective function given in Equation \ref{eq:objective} enables backpropagation of the objective gradients to the network's weights. The training procedure is then straightforwardly framed as an optimisation task computable e.g. via conjugate gradient or stochastic gradient descent. 

\subsection{Approximating the Reward Structure}
\label{sec:RewardStructure}
Due to the dimensionality and size of the state space in many real world applications, the reward structure can not be observed explicitly for every state. In these cases state rewards are not modelled directly per state, but the reward structure is restricted by imposing that states with similar features, $x$, should have similar rewards. To this end, function approximation is used in order to regress the feature representation onto a real valued reward using a mapping $g : \mathbb{R}^N \rightarrow \mathbb{R}$, with $N$ being the dimensionality of the feature space such that
\begin{equation}
r = g(f,\theta).
\end{equation}
A feature representation, $f$, is usually hand-crafted based on preprocessing such as segmentation and manually defined distance metrics, but can be learned based on the proposed framework - as shown in section \ref{sec:cnn}. Furthermore, the application of feature based function approximation enables easier generalisation and transfer of models.

The choice of model used for function approximation has a dramatic impact on the ability of the algorithm to capture relationship between the state feature vector $f$ and user preference. Commonly, the mapping from state to reward is simply a weighted linear combination of feature values
\begin{equation}
g(f,\theta) = \theta^\top f.
\end{equation}
This choice, while appropriate in some scenarios, is suboptimal if the true reward can not be accurately approximated by a linear model. In order to alleviate this limitation \cite{choi2013bayesian} extend the linear model by introducing a mapping $\Phi : \mathbb{R}^N \rightarrow \{0,1\}^N$ such that 
\begin{equation}
g(f,\theta,\Phi) = \theta^\top \Phi(f).
\end{equation}
Here $\Phi$ denotes a set of composite features which are jointly learned as part of the objective function. These composites are assumed to be the logical conjunctions of the predefined, atomic features $f$. Due to the nature of the features used the representational power of this approach is limited to the family of piecewise constant functions. 

In contrast, \cite{levine2011nonlinear} employ a Gaussian Processes (GP) framework to capture the potentially unbounded complexity of any underlying reward structure. The set of expert demonstrations $\mathcal{D}$ is used in this context to identify an active set of GP support points, $X_u$, and associated rewards $u$. The mean function is then used to represent the individual reward at a state described by $f$

\begin{equation}
g(f,\theta,\mathcal{X}_u,u) = K^\top_{f,u} K^{-1}_{u,u} u.
\end{equation}
Here $K_{f,u}$ denotes the covariance of the reward at $f$ with the active set reward values $u$ located at $X_u$ and $K_{u,u}$ denotes the covariance matrix of the rewards in the active set computed via a covariance function $k_\theta(f_i,f_j)$ with hyperparameters $\theta$.

Nevertheless, a significant drawback of the GPIRL approach is a computational complexity proportional to the number of demonstrations and the size of the active set of inducing points, which in turn depends on the reward complexity. While the modelling of complex, nonlinear reward structures in problems with large state spaces is theoretically feasible for the GPIRL approach, the cardinality of the active set will quickly become unwieldy, putting GPIRL at a significant computational disadvantage or, worse, rendering it entirely intractable. These shortcomings are remedied when using deep parametric architectures for reward function approximation while keeping the accuracy of nonlinear function approximation, as outlined in the next section.

\section{Reward Function Approximation with Deep Architectures}
We argue that IRL algorithms scalable to MDPs with large feature spaces require models, which are able to efficiently represent complex, nonlinear reward structures. In this context, deep architectures are a natural choice as they explicitly exploit the depth-breadth trade-off \cite{bengio2007scaling} and increase representational capacity by reusing the computations of earlier nodes in the following layers.

\begin{figure}
	\centering
	\includegraphics[height=1.6in]{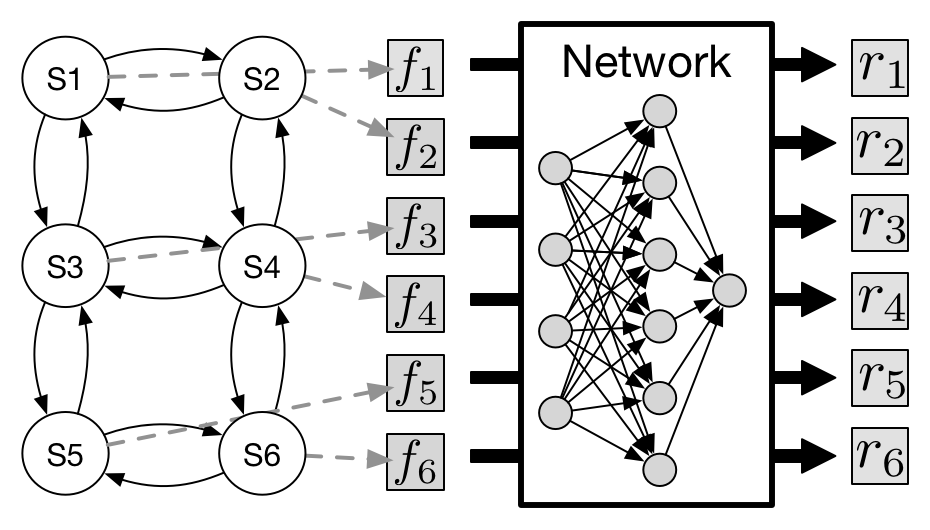}
	\caption{Schema for Neural Network based reward function approximation based on the feature representation of MDP states}
	\label{fig:reward_approx}
\end{figure}

For the remainder of the paper, we consider a network architecture which accepts as input state features $x$, maps these to state reward $r$ and is governed by the network parameters $\theta_{1,2,..n}$. 
In the context of Section~\ref{sec:RewardStructure}, the state reward is therefore obtained as
\begin{eqnarray}
r & \approx & g(f,\theta_1,\theta_2,...,\theta_n)\\
&=& g_1(g_2(...(g_n(f,\theta_n),...),\theta_2),\theta_1).
\end{eqnarray}
%
While many choices exist for the individual building blocks of a deep architecture, it has been shown that a sufficiently large NN with as little as two layers and sigmoid activation functions can represent any binary function \cite{hassoun1995fundamentals} or any piecewise-linear function \cite{hornik1989multilayer} and can therefore be regarded as a \emph{universal approximator}. While this holds true in theory, it can be far more computationally practicable to extend the depth of the network structure and reduce the number of required computations in doing so \cite{bengio2009learning}.

Importantly, in applying backpropagation, NNs also lend themselves naturally to training in the maximum entropy IRL framework and the network structure can be adapted to suit individual tasks without complicating or even invalidating the main IRL learning mechanism. 
In the DeepIRL framework proposed here the full range of architecture choices thus becomes available. Different problem domains can utilise different network architectures as e.g. convolutional layers can remove the dependency on handcrafted spatial features. Furthermore, it is straightforward to show that the linear maximum entropy IRL approach proposed in \cite{ziebart2008maximum} can be seen as a simplification of the more general deep approach and can be created by applying the rules of back-propagation to a network with a single linear output connected to all inputs with zero bias term.

While the common NN architectures for whole-image classification regress to fixed size outputs, the applied FCNNs result in an output with equivalent spatial dimensionality and by padding data correspondingly we realise reward maps of the same size as our input. It is to note here that padding is not the only possibility and deconvolutions as applied by \cite{LongSD14} can also transform and reshape to create equally sized model output.



\subsection{Training Procedure}
\label{subsec:learning}

The task of solving the IRL problem can be framed in the context of Bayesian inference as MAP estimation, maximizing the joint posterior distribution of observing expert demonstrations,  $\mathcal{D}$, under a given reward structure and of the model parameters $\theta$. 
\begin{equation}
\mathcal{L(\theta)} = \log P(\mathcal{D},\theta|r) = \underbrace{\log P(\mathcal{D}|r)}_{\mathcal{L_D}} + \underbrace{\log P(\theta)}_{\mathcal{L_\theta}}.
\end{equation}

This joint log likelihood is differentiable with respect to the parameters $\theta$ of a linear reward model, which allows the application of gradient descent methods \cite{snyman2005practical}. 
We extend this benefit with the adaptation of Maximum Entropy for neural networks as presented in $\mathcal{L_D}$ of Equation \ref{eq:objective} by separating into the gradient of the loss with respect to the rewards $r$ and the gradient of the reward with respect to the network's weights obtained via backpropagation.

The complete gradient is given by the sum of the gradients with respect to $\theta$ of the data term $\mathcal{L_D}$ and a weight decay term as model regulariser $\mathcal{L_\theta}$
\begin{equation}\label{eq:objectivefun}
\frac{\partial \mathcal{L}}{\partial \theta} = \frac{\partial \mathcal{L_D}}{\partial \theta} + \frac{\partial \mathcal{L_\theta}}{\partial \theta}.
\end{equation}

The earlier mentioned separation of derivatives in the gradient of the data term is shown in equation
\ref{eq:objective}

\begin{eqnarray}\label{eq:objective}
\frac{\partial \mathcal{L_D}}{\partial \theta} &=& \frac{ \partial \mathcal{L_D} }{ \partial r } \cdot \frac{ \partial r }{ \partial \theta }\\
&=& (\mu_{\mathcal{D}} - \mathbb{E} [\mu]) \cdot \frac{\partial}{\partial \theta} g(f,\theta) , 
\end{eqnarray}
where  $r = g(f,\theta))$. As shown in \cite{ziebart2008maximum}, the gradient of the expert demonstration term $\mathcal{L_D}$ with respect to the model parameters of a linear function is equal to the difference in feature counts along the trajectories. For higher level models this gradient can be split into the derivative with respect to the reward $r$ and the derivative of the reward with respect to the model parameters which in case of a neural network is obtained via backpropagation. The derivative of the Maximum Entropy objective with respect to the reward equals the difference in state visitation counts between solutions given by the expert demonstrations and the expected visitation counts for the learned systems trajectory distribution in \ref{eq:expected}.

\begin{equation}\label{eq:expected}
\mathbb{E} [\mu] = \sum_{\varsigma : \{s,a\} \in \varsigma} P(\varsigma | r)
\end{equation}
Computation of $\mathbb{E} [\mu]$ usually involves summation over exponentially many possible trajectories. A more effective algorithm based on dynamic programming which computes this quantity in polynomial-time can be found in \cite{ziebart2008maximum,kitani2012activity}. Subsequently, the effective computation of the  gradient $\frac{\partial \mathcal{L_D}}{\partial \theta}$ involves first computing the difference in visitation counts using this algorithm and then passing this as an error signal through the network using back-propagation. 


The complete proposed method is described by Algorithm \ref{alg:medirl}, with the loss and gradient derivation in lines 6 and 7 given by the linear Maximum Entropy formulation. The expert's state action frequencies $\mu_D^a$, which are needed for the calculation of the loss are summed over the actions to compute the expert state frequencies $\mu_D = \sum \limits_{a=1}^A \mu_D^a$ .

\begin{algorithm}[t]
	\caption{Maximum Entropy Deep IRL}  
		\begin{algorithmic}[1]\label{alg:medirl} 
		    \begin{spacing}{1.5}
			\REQUIRE $\mu_D^a, f, S, A, T, \gamma$
			\ENSURE $\mathrm{optimal ~ weights}~ \theta^*$ 
			\vspace{1mm}
			\STATE{$\theta^1 = \mathrm{initialise\_weights()}$} \\
			\vspace{3mm}
			{\textcolor{myblue}{\textbf{Iterative model refinement}}}\\
			\FOR{n = 1 : N} 
			\STATE{$r^n = \mathrm{nn\_forward}(f,\theta^n)$ } \\
			\vspace{3mm}
			{ \textcolor{myblue}{{\textbf{Solution of MDP with current reward}}}}\\
			\STATE{$\pi^n = \mathrm{approx\_value\_iteration}(r^n,S,A,T,\gamma)~~~~~~~~~$} \\ 
			\STATE{$\mathbb{E} [\mu^n] = \mathrm{propagate\_policy}(\pi^n,S,A,T)~~~~$}  \\ 
			\vspace{3mm}
			{ \textcolor{myblue}{{\textbf{Determine Maximum Entropy loss and gradients}}}}\\
			\STATE{$\mathcal{L}_D^n = \log(\pi^n) \times \mu_D^a$ }\\
			\STATE{$\frac{\partial \mathcal{L}_D^n}{\partial r^n}  = \mu_D - \mathbb{E} [\mu^n]$ }\\
			\vspace{3mm}
			{ \textcolor{myblue}{{\textbf{Compute network gradients}}}}\\
			\STATE{$\frac{\partial \mathcal{L}_D^n}{\partial \theta_D^n} = \mathrm{nn\_backprop}(f, \theta^n, \frac{\partial \mathcal{L}_D^n}{\partial r^n}) ~~$}  \\ 
			\STATE{$\theta^{n+1} = \mathrm{update\_weights}(\theta^n, \frac{\partial \mathcal{L}_D^n}{\partial \theta_D^n} ) $}
			\ENDFOR
			\end{spacing}
		\end{algorithmic}
	
\end{algorithm}

Lines 4 and 5 are explained in detail in the algorithms \ref{alg:backward} and \ref{alg:forward} respectively, and are adapted from \cite{kitani2012activity}. Algorithm \ref{alg:backward} determines the policy given the current reward model via iterative update of the state-action value function, while algorithm \ref{alg:forward} determines the expected state visiting frequencies by probabilistically traversing the MDP given the current policy. Additional indices representing the iteration of the main algorithm were omitted in these subscripts in favour of readability.

The presented algorithm is applied to train FCNNs based on the loss derivatives for all states at once. As each of the final state-wise rewards is influenced by its corresponding area in the original state space -- its receptive field, training with the summed loss over the whole scene is equivalent to a stochastic gradient formulation with all receptive fields addressed in a minibatch. This formulation is computationally more efficient than separate computation per field, since these fields overlap as soon as the width of our convolutional filters exceeds one \cite{LongSD14}.

\begin{algorithm}[t]
	\caption{Approximate Value Iteration}  
	\begin{spacing}{1.5}
		\begin{algorithmic}[1]\label{alg:backward} 
			\STATE{$V(s) = -\infty $} 
			\REPEAT
		    \STATE{$V_t=V; ~~V(s_{goal}) = 0$ }
		    \STATE{$Q (s,a) = r(s,a) + E_{T(s,a,s')}[V(s')]$ }
		    \STATE{$V=softmax_a ~ Q_i(s,a)$ }
			\UNTIL{$max_s(V(s)-V_t(s)) < \epsilon$}
			\STATE{$\pi(a|s) = e^{Q(s,a)-V(s)}$ }
			
		\end{algorithmic}
	\end{spacing}
\end{algorithm}

\begin{algorithm}[t]
	\caption{Policy Propagation}  
	\begin{spacing}{1.5}
		\begin{algorithmic}[1]\label{alg:forward} 

			\STATE{$\mathbb{E}_1 [\mu(s_{start})] = 1$} 
			\FOR{i = 1 : N} 
            \STATE{$\mathbb{E}_i [\mu(s_{goal})] = 0 $}
            \STATE{$\mathbb{E}_{i+1} [\mu(s)] = \sum_{s',a} T(s,a,s')~ \pi(a|s')~ \mathbb{E}_{i} [\mu(s')] $}
            \ENDFOR
            \STATE{$\mathbb{E} [\mu(s)] = \sum_i \mathbb{E}_i [\mu(s)]$}
			
		\end{algorithmic}
	\end{spacing}
\end{algorithm}

\section{Experiments}
\label{experiments}
We assess the performance of DeepIRL two benchmark tasks against current state-of-the-art approaches
: GPIRL~\cite{levine2011nonlinear}, NPB-FIRL~\cite{choi2013bayesian} and the original MaxEnt~\cite{ziebart2008maximum} to illustrate the necessity of non-linear function approximation.

All tests are run multiple times on training and transfer scenarios for the different settings, while learning is performed based on synthetically generated stochastic demonstrations based on the optimal policy to evaluate performance on suboptimal example sets.
This is achieved by providing a number of demonstrations sampled from the optimal policy based on the true reward structure, but including 30\% of random actions.

In our experiments, we employ a FCNN with two hidden layers and rectified linear units as function approximator between state feature representation and reward.
This rather shallow networks structure suffices for the application based on strongly simplified toy benchmarks. However, the whole framework can be utilised for training networks of arbitrary capacity. All experiments except for the spatial feature learning in section \ref{sec:cnn} are based on filters of width one to focus on direct evaluation against the other algorithms, which are in their current form limited to the features of each state for reward approximation. Wider filters as applied for spatial feature learning are used to evaluate the performance on raw inputs without manual feature design. 
For these benchmarks, we apply AdaGrad \cite{duchi2011adaptive}, an approach for stochastic gradient descent with per parameter adaptive learning rates. Significant parts of the neural network implementation are based on MatConvNet \cite{vedaldi2014mcn}.

In line with related works, we use \emph{expected value difference} as principal metric of evaluation. It is a measure of the sub-optimality of the learned policy under the true reward. The score represents the difference between the value function obtained for the optimal policy given the true reward structure and the value function obtained for the optimal policy based on the learned reward model. Additionally to the evaluation on each specific training scenario, the trained models are evaluated on a number of randomly generated test environments. The test on these \emph{transfer} examples serves to analyse each algorithm's ability to generalise to the true reward structure without over-fitting.

\subsection{Objectworld Benchmark}
\label{objectworld}

The \emph{Objectworld} scenario \cite{levine2011nonlinear} consists of a map of $M \times M$ states for $M = 32$ where possible actions include motions in all four directions as well as staying in place. Two different sets of state features are implemented based on randomly placed colours to evaluate the algorithms. For the continuous set $x \in \mathbb{R}^C$. Each feature dimension describes the minimum distance to an object of one of $C$ colours. Building on the continuous representation the discrete set includes $C\times M$ binary features, where each dimension indicates whether an object of a given colour is closer than a threshold $d \in \{1, ..., M\}$.

\begin{figure}[h]
	\centering
	\includegraphics[height=3cm]{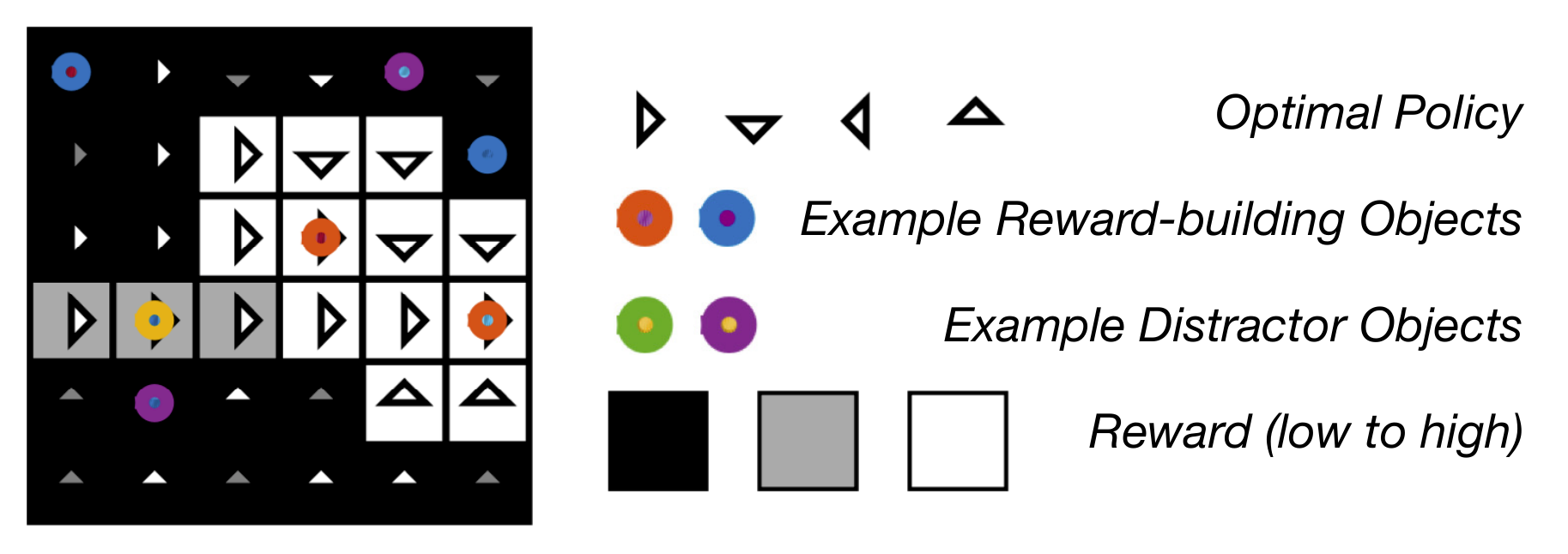} 
	\caption{Objectworld benchmark. The true reward is displayed by the brightness of each cell and based on the surrounding object configuration. Only a subset of colors influences the reward, while the others serve as distracting features.
	}
	\label{fig:reward_object2}
\end{figure}

The reward is positive for cells which are both within the distance 3 of color 1 and distance 2 of color 2, negative if only within distance 3 of color 1 and zero otherwise. This is illustrated for a small subset of the state space in Figure \ref{fig:reward_object2}. 

In line with common benchmarking procedures, we evaluated the algorithms with a set number of features and increasing demonstrations. Additionally, the learned reward functions are deployed on randomly generated transfer scenarios to uncover any overfitting to the training data. 

\begin{figure}[h]
	\centering
	\begin{tabular}{c c}
		\includegraphics[width=\fisb,trim = 3cm 8.1cm 5.1cm 0cm, clip]{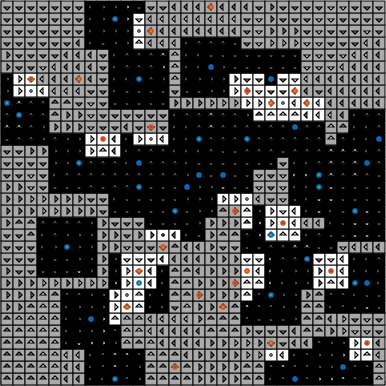} &
		\includegraphics[width=\fisb,trim = 3cm 8.1cm 5.1cm 0cm, clip]{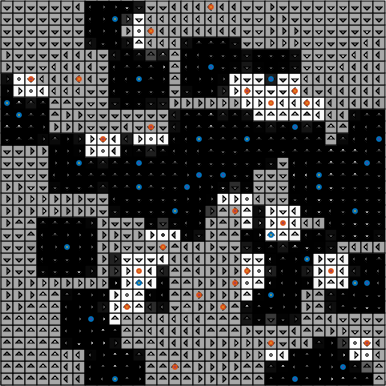} \\
		Groundtruth & DeepIRL \\
		\includegraphics[width=\fisb,trim = 3cm 8.1cm 5.1cm 0cm, clip]{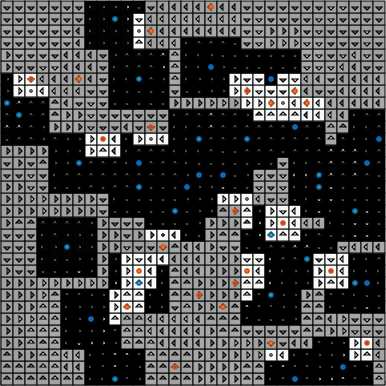} &
		\includegraphics[width=\fisb,trim = 3cm 8.1cm 5.1cm 0cm, clip]{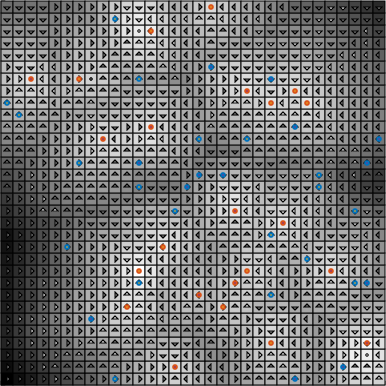} \\
		GPIRL & MaxEnt \\
		
	\end{tabular}
	\caption{Reward reconstruction sample in Objectworld benchmark provided $N=64$ examples and $C=2$ colours with continuous features. White - high reward; black - low reward.}
	\label{fig:reward_obj}
\end{figure}

\begin{figure} [ht]
	\centering
	\begin{tabular}{c c}
		
		\includegraphics[width=\fis]{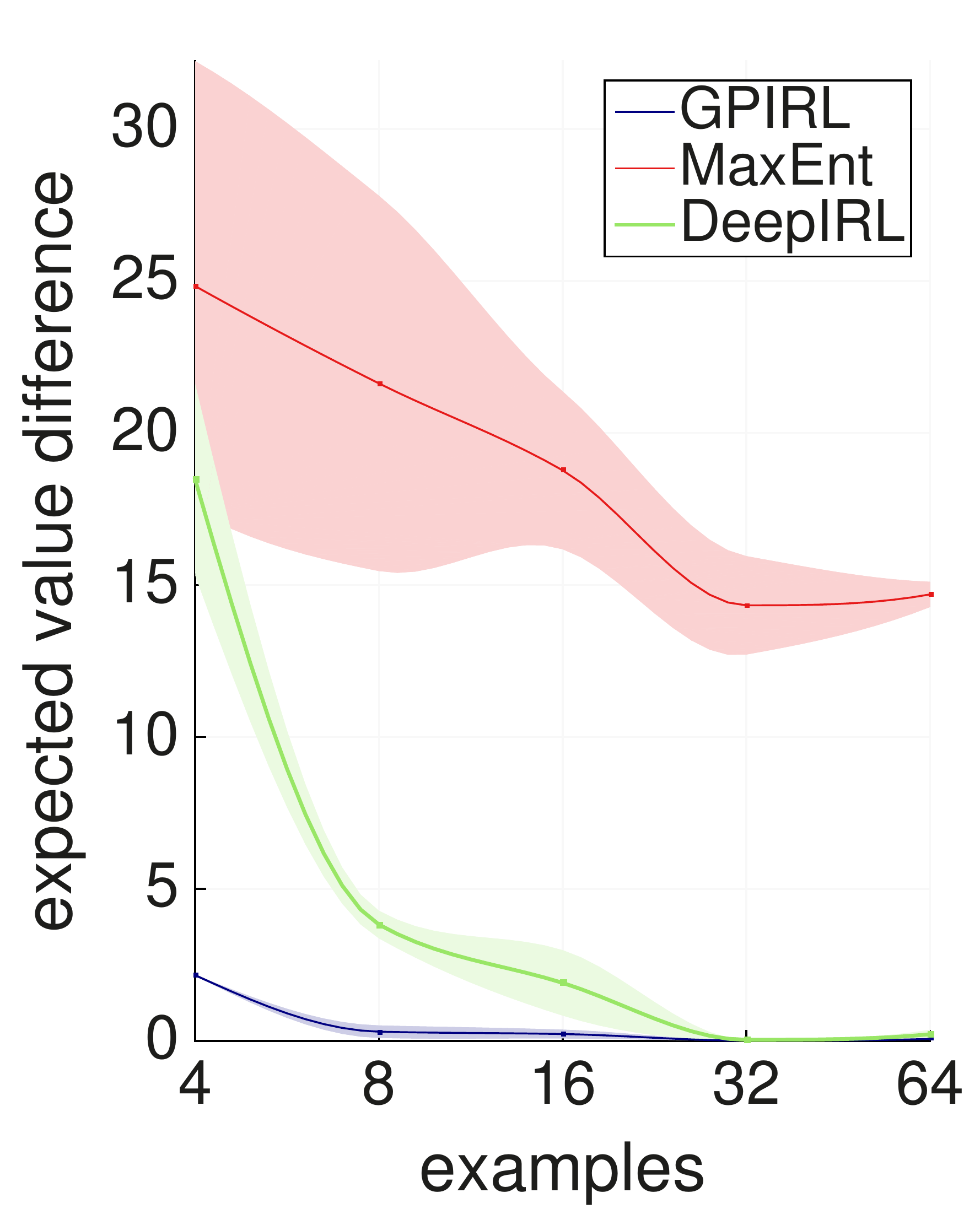} &
		~ 
		\includegraphics[width=\fis]{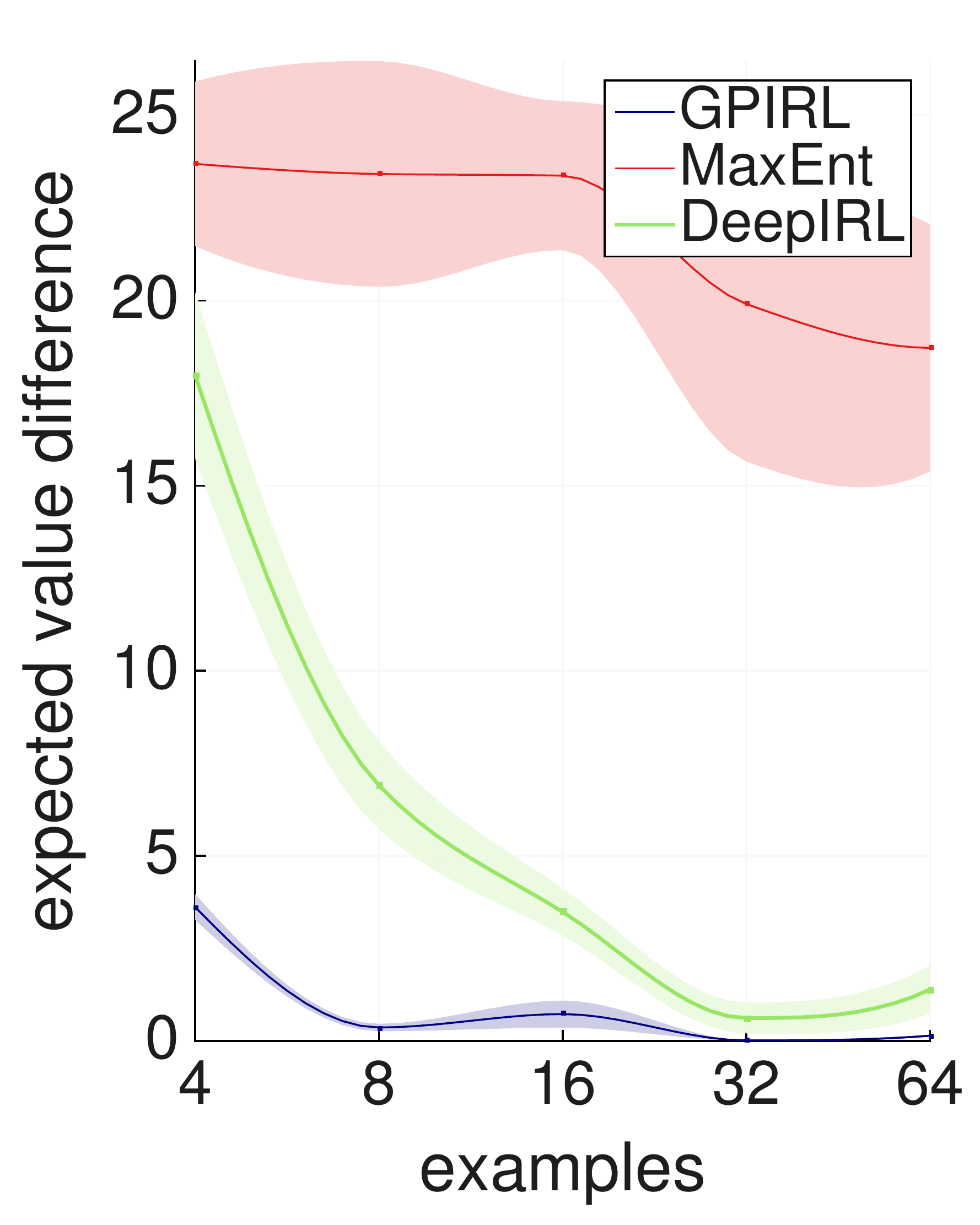}
		\\
		a) & b)
		\\
		\includegraphics[width=\fis]{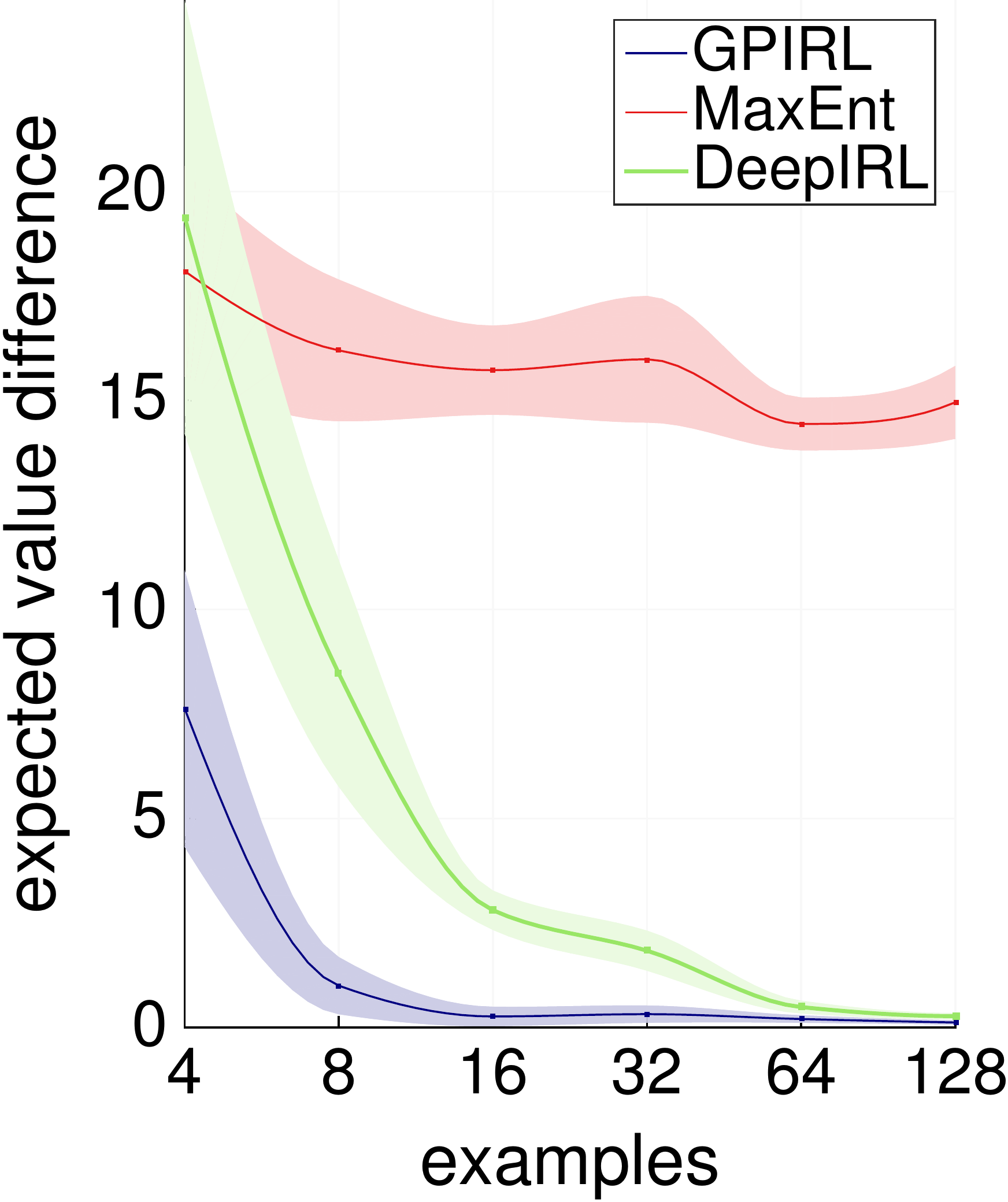} &
		~
		\includegraphics[width=\fis]{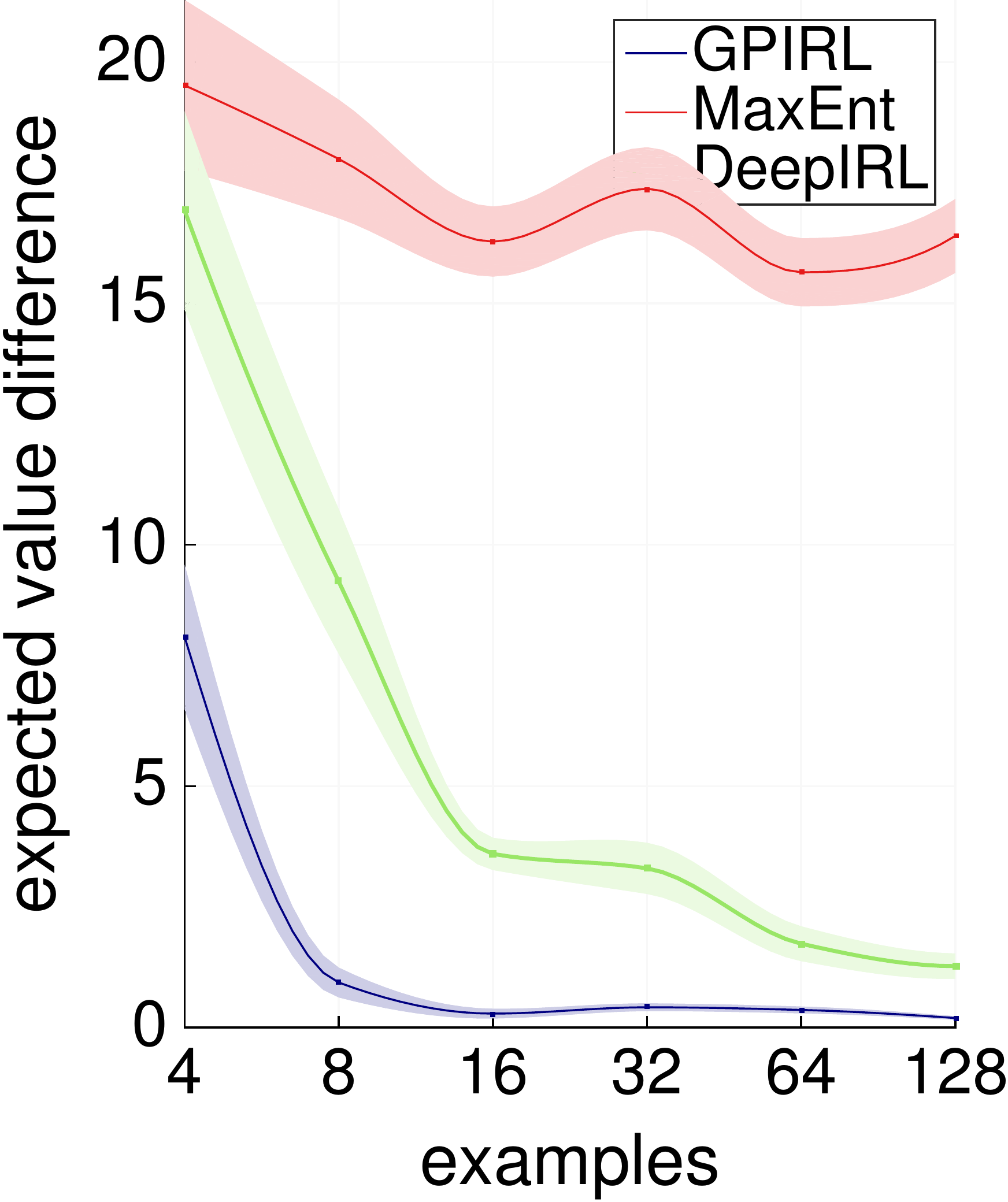}
		\\
		c) & d)
	\end{tabular}
	\caption{Objectworld benchmark. From top left to bottom right: expected value difference (EVD) with $C=2$ colours and varying number of demonstrations $N$ for training a) and transfer case b) with continuous and subsequently with discrete features in c) \& d) ; 
		As the number of demonstrations grows DeepIRL is able to quickly match performance of GPIRL on the task.}\label{fig:experiment1}
\end{figure}

While the original MaxEnt is unable to capture the nonlinear reward structure well, both DeepIRL and GPIRL provide significantly better approximations as represented in Figure \ref{fig:reward_obj}. 
Evaluation of NPB-FIRL on this benchmark was done in \cite{choi2013bayesian} where it showed a similar level of performance as GPIRL.
GPIRL generates a good model already with few data points whereas DeepIRL achieves commensurate performance when increasing the number of available expert demonstrations. The same behaviour is exhibited when using both continuous and discrete state features (Fig.~\ref{fig:experiment1}). The requirement for more training data will be rendered unimportant in robot applications based on autonomous data acquisition, while enforcing the lower algorithmic complexity as dominant advantage of the parametric approach.

Additional tests are performed with increased number of distractor features to evaluate each approach's overfitting tendency. The corresponding figures are left out due to limited space. Both DeepIRL and GPIRL show robustness to distractor variables, though DeepIRL shows minimally bigger signs of overfitting as the number of distractor variables is increased. This is due to the NN's capacity being brought to bear on the increasing noise introduced by the distractors and will be addressed in future work with additional regularisation methods, such as Dropout \cite{hinton2012dropout} and ensemble methods.

\subsection{Binaryworld Benchmark}
In order to test the ability of all approaches to successfully approximate more complex reward structures, the \emph{Binaryworld} benchmark is presented. This test scenario is similar to \emph{Objectworld}, but in this problem every state is randomly assigned one of two colours (blue or red). The feature vector for each state consequently consists of a binary vector of length 9, encoding the colour of each cell in its 3x3 neighbourhood. The true reward structure for a particular state is fully determined by the \emph{number} of blue states in its local neighbourhood. It is positive if exactly four out of nine neighbouring states are blue, negative if exactly five are blue and zero otherwise. The main difference compared to the Objectworld scenario is that a single feature value does not carry much weight, but rather that higher-order relationships amongst the features determine the reward. 

Since the reward depends on a higher representation for the basic features - that is to say the number of specific features - such case is arguably more challenging than the original Objectworld experiment and a good performance on this benchmark implies the algorithm's ability to learn and capture this complex relationship.

\begin{figure}
	\centering
	\begin{tabular}{c c}
		\includegraphics[width=\fis]{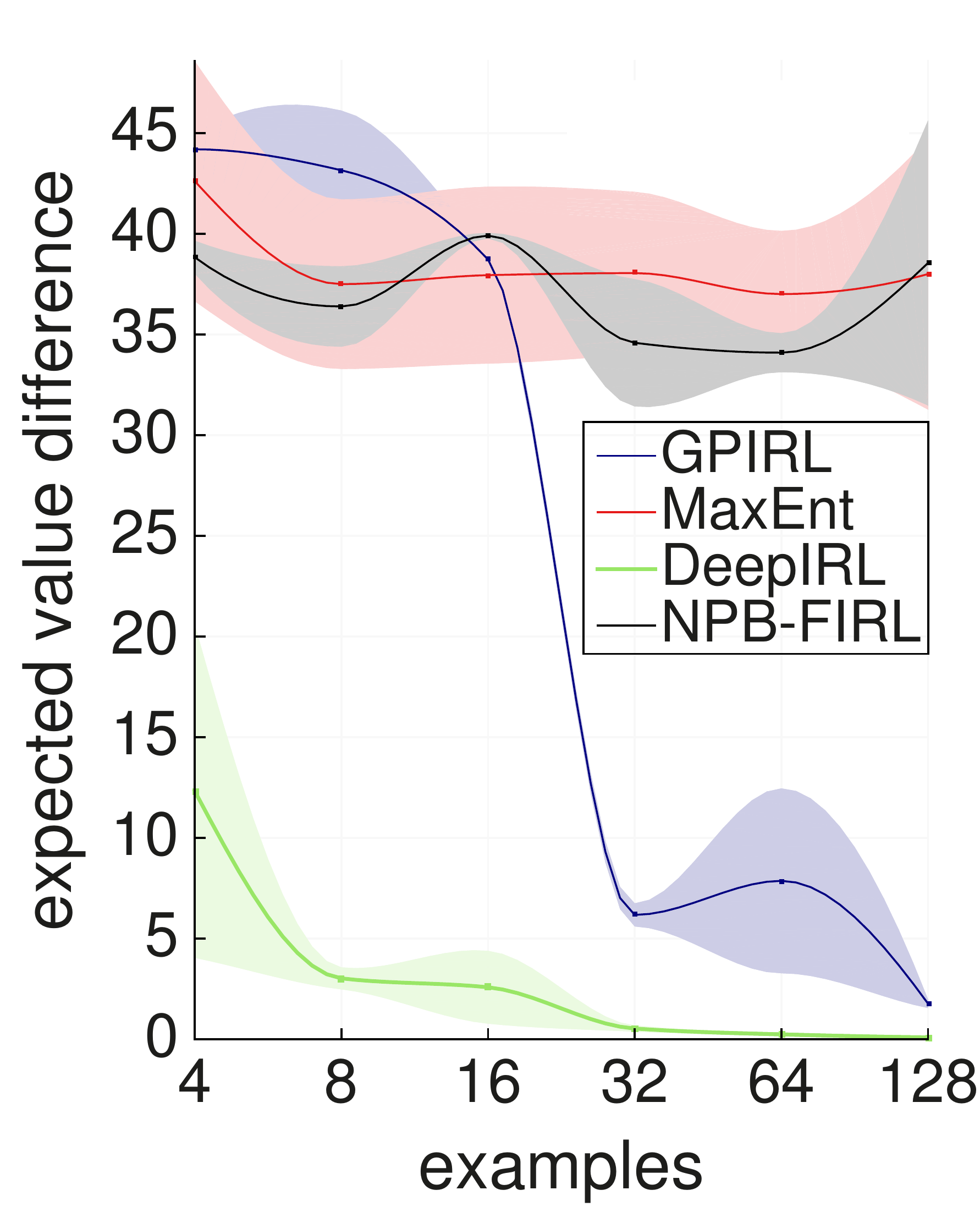} & 
		\includegraphics[width=\fis]{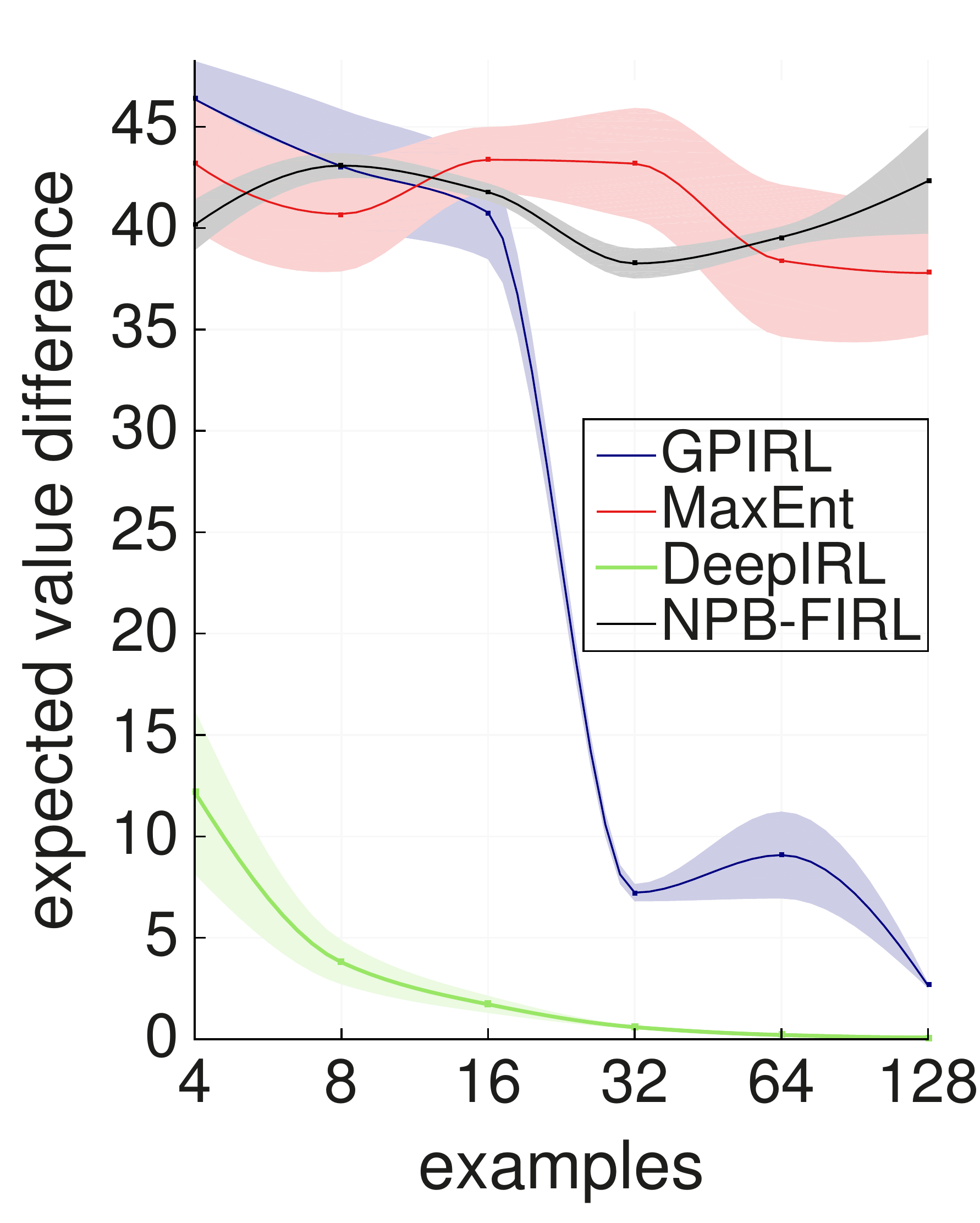}
	\end{tabular}
	\caption{Value differences observed in the Binaryworld benchmark for GPIRL, MaxEnt and DeepIRL for the training scenario (left) and the transfer task (right).}
	\label{fig:experiment2}
\end{figure}

The performance of DeepIRL compared to GPIRL, linear MaxEnt and NPB-FIRL is depicted in Fig.~\ref{fig:experiment2}. In this increasingly more complex scenario, DeepIRL is able to learn the higher-order dependencies between features, whereas GPIRL struggles as the inherent kernel measure can not correctly relate the reward of different examples with similarity in their state features. 
GPIRL needs a larger number of demonstrations to achieve good performance and to determine an accurate estimate on the reward for all $2^9$ possible feature combinations.

Perhaps surprising is the comparatively low performance of the NPB-FIRL algorithm. This can be explained by the limitations of this framework. The true reward in this scenario can not be efficiently described by the logical conjunctions used. In fact, it would require $2^9$ different logical conjunctions, each capturing all possible combinations of features, to accurately model the reward in this framework.

\begin{figure} [h]
	\centering
	\begin{tabular}{c c}
		\includegraphics[width=\fisb,trim = 3cm 8.1cm 5.1cm 0cm, clip]{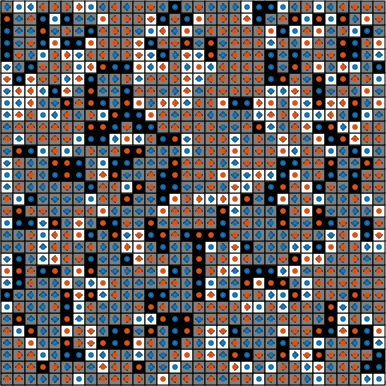} & 
		\includegraphics[width=\fisb,trim = 3cm 8.1cm 5.1cm 0cm, clip]{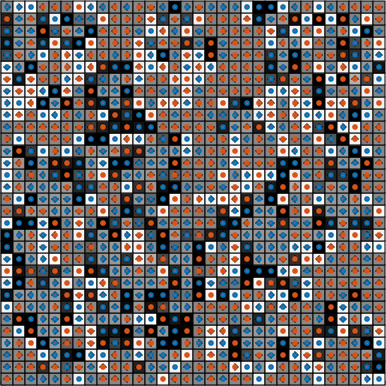} \\
		Groundtruth & DeepIRL \\
		\includegraphics[width=\fisb,trim = 3cm 8.1cm 5.1cm 0cm, clip]{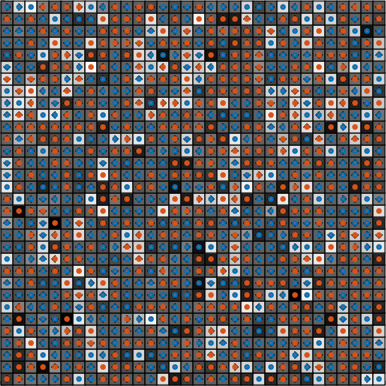} &
		\includegraphics[width=\fisb,trim = 3cm 8.1cm 5.1cm 0cm, clip]{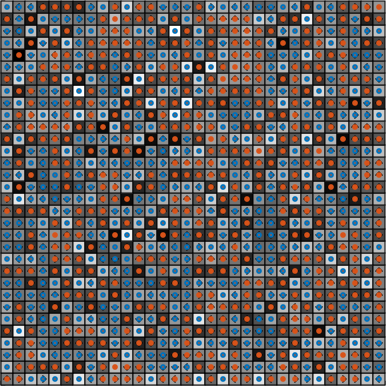} \\
		GPIRL & MaxEnt
	\end{tabular}
	\caption{Reward reconstruction sample for the Binaryworld benchmark provided $N=128$ demonstrations. White - high reward; black - low reward.}
	\label{fig:reward_bin}
\end{figure}

Fig.~\ref{fig:reward_bin} shows the reconstruction of the reward structures estimated by DeepIRL, MaxEnt and GPIRL. While GPIRL was able to reconstruct the correct reward for some of the states having features it has encountered before it provides inaccurate rewards for states which were never encountered. It produces an overall too smooth reward function due to assumptions and priors in the GP approximation. On the other hand, DeepIRL is able to reconstruct it with high accuracy demonstrating the ability to effectively learn the highly-varying structure of the underlying function.

\subsection{Spatial Feature Learning} \label{sec:cnn}

\begin{figure}[h]
	\centering
	\begin{tabular}{c c}
		\includegraphics[width=\fis]{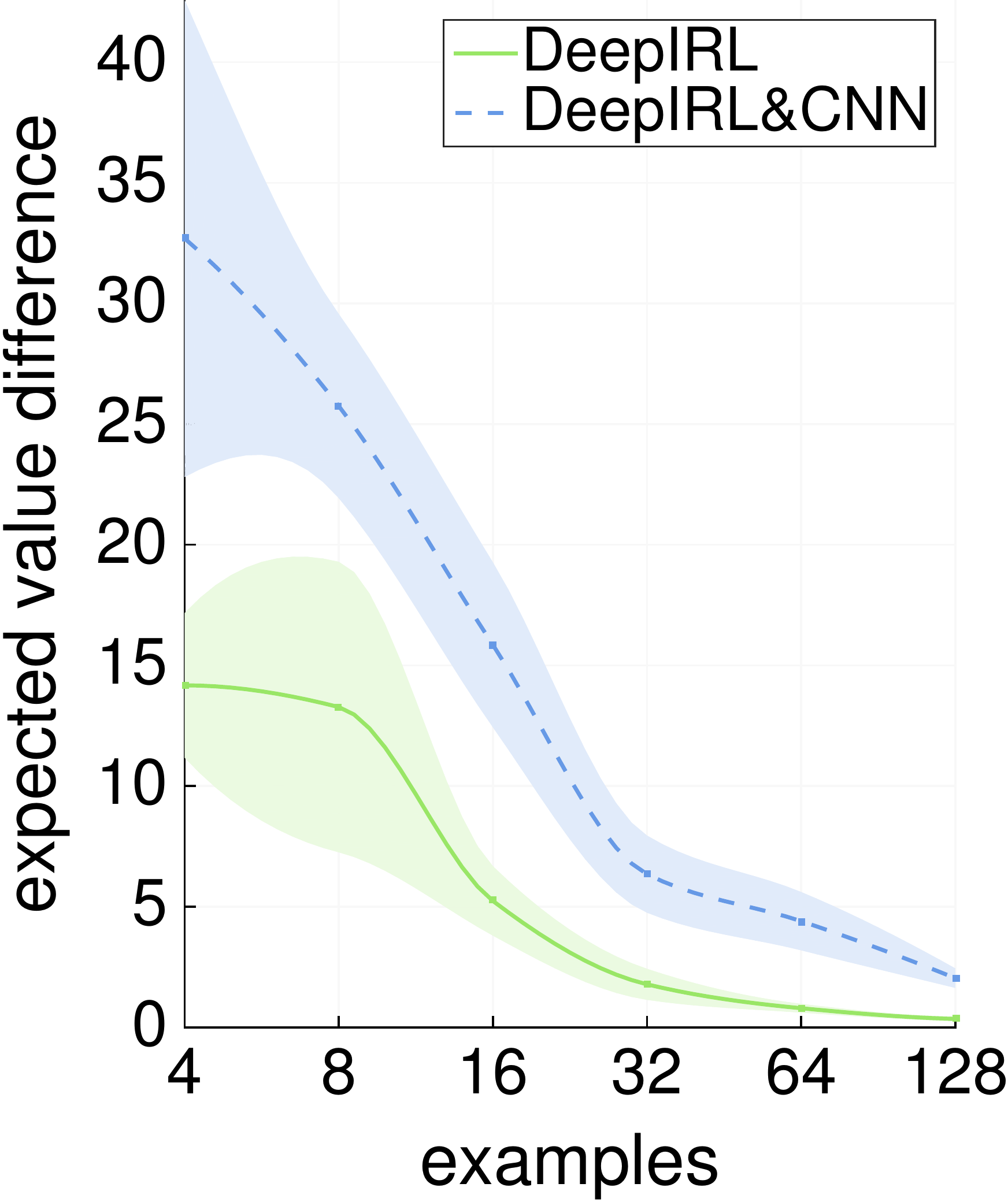} & 
		\includegraphics[width=\fis]{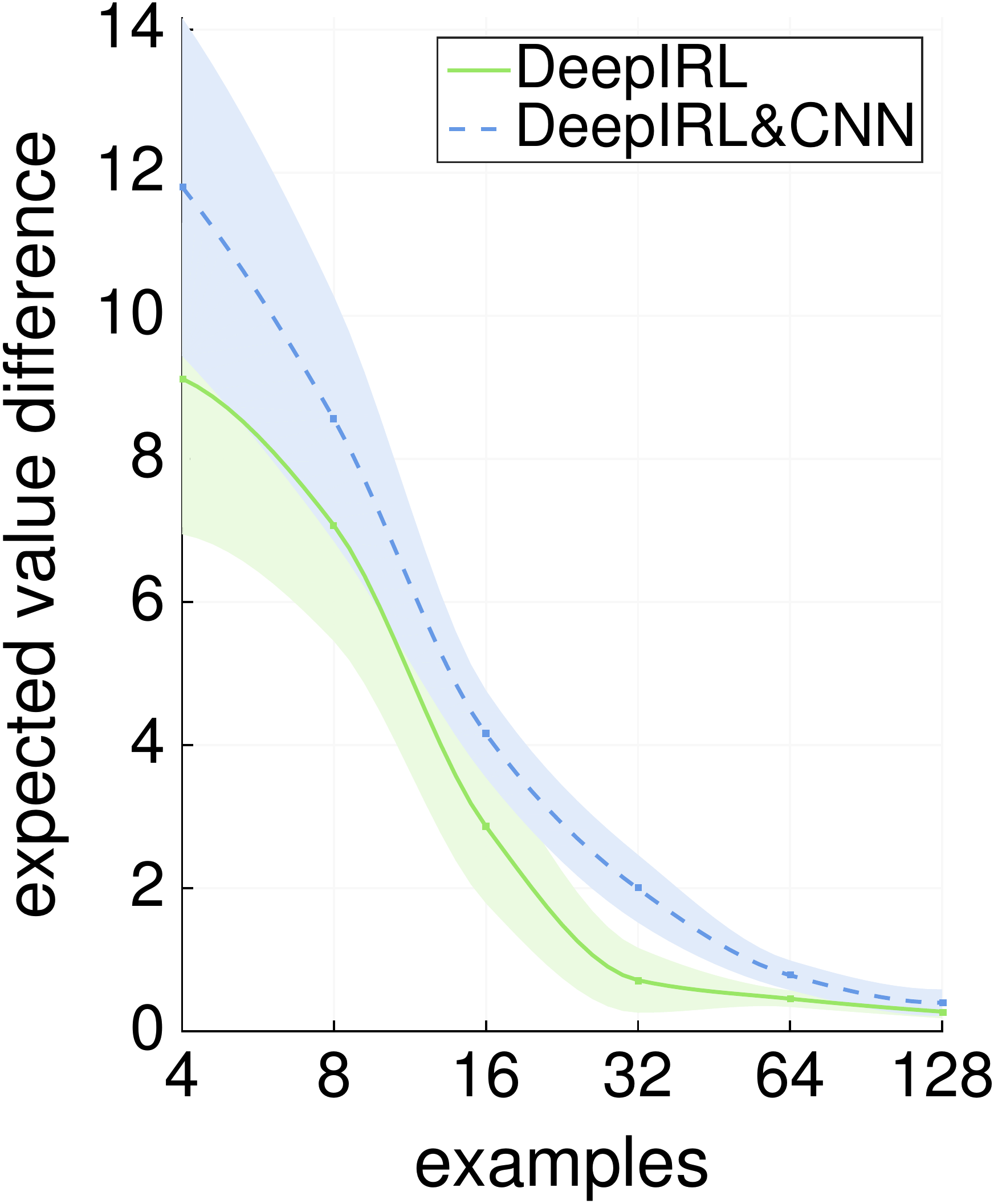} \\
		Objectworld & Binaryworld
	\end{tabular}
	\caption{Application of convolutional layers for spatial feature learning. Spatial feature learning quickly converges to performance with optimally designed features.}
	\label{fig:cnn}
\end{figure}

While the earlier benchmarks visualise performance compared to current algorithms in the context of precomputed features, the approach can be extended via the use of wider filters to eliminate the requirement of preprocessing or manual design of features. Figure \ref{fig:cnn} represents the results for both earlier benchmarks, but instead of using the earlier described feature representations, the FCNN builds the reward based on the raw input representation, which for each state only includes the availability of each specific object at that specific state. All spatial information is derived based on the convolutional filters. Based on the simplicity of the benchmarks, we employed a five layer approach with 3x3 convolutional kernels in the first two layers. 
By increasing the depth of the network and include convolutional filters, we add enough capacity to enable the learning of features as well as their combination into the reward function in the same architecture and process.

Due to the increasing number of parameters, the approach requires additional training data to perform at equal accuracy but with increasing number of expert samples converges towards the performance with predefined features. Since the given features in these simplified toy problems are optimal and the true reward is directly calculated on their basis, automatically learned features cannot exceed the performance. However, in real-world scenarios, the compression of raw data - such as images - to feature representations leads to information loss and the learning of task-relevant features gains even more importance.

\section{Conclusion and Future Work}
This paper presents Maximum Entropy Deep IRL, a framework exploiting FCNNs for reward structure approximation in Inverse Reinforcement Learning. Neural networks lend themselves naturally to this task as they combine representational power with computational efficiency compared to state-of-the-art methods. Unlike prior art in this domain DeepIRL can therefore be applied in cases where complex reward structures need to be modelled for large state spaces. Moreover, training can be achieved effectively and efficiently within the popular Maximum Entropy IRL framework. A further advantage of DeepIRL lies in its versatility. Custom network architectures and types can be developed for any given task while exploiting the same cost function in training. This is expressed in section \ref{sec:cnn}, where convolutional filters are applied to eliminate the need of manual feature design.

Our experiments show that DeepIRL's performance is commensurate to the state-of-the-art on a common benchmark. 
While exhibiting slightly increased requirements regarding training data in this benchmark, a principal strength of the approach lies in its algorithmic complexity independent of the number of demonstrations samples. Therefore, it is particularly well-suited for life-long learning scenarios in the context of robotics, which inherently provide sufficient amounts of training data.
We also provide an alternative evaluation on a new benchmark with a significantly more complex reward structure, where DeepIRL significantly outperforms the current state-of-the-art and proves its strong capability in modeling the interaction between features. Furthermore, we extend the approach to wider filters in order to eliminate the dependency on precomputed features and to emphasise the adaptability of framing IRL in the context of deep learning.

In future work we will explore the benefits of autoencoder-style pretraining to reduce the increased demand of expert demonstrations when employing wider convolutional filters. Especially when based on more complex inputs such as raw image data, the easily available unsupervised training data will help to learn features which then only need to be refined during the supervised IRL-based training phase.
Due to the variety of existing work on FCNN architectures mentioned in section \ref{introduction}, we expect to be able to benefit from applying more complex networks for real life problems, such as the skipping architecture by \cite{LongSD14}, which enables the concatenation of fine structural information alongside with coarser higher level features in the last regression layer to improve overall performance in evaluating features of multiple scales. Furthermore, other methods for optimising demonstration data likelihood such as given by \cite{babes2011apprenticeship} will be evaluated.

\bibliography{main}
\bibliographystyle{icml2016}

\end{document}